\documentclass[11pt]{article}

\usepackage[utf8]{inputenc}
\usepackage[T1]{fontenc}
\usepackage{amsmath,amssymb}
\usepackage{graphicx}
\usepackage{booktabs}
\usepackage{array}
\usepackage{cite}
\usepackage{url}
\usepackage{geometry}
\graphicspath{{fig/}}

\title{Prediction-space knowledge markets for communication-efficient federated learning on multimedia tasks}

\author{Du Wenzhang\\
Department of Computer Engineering\\
Mahanakorn University of Technology\\
International College (MUTIC)\\
Bangkok, Thailand\\
\texttt{dqswordman@gmail.com}}

\begin{document}

\maketitle

\begin{abstract}
Federated learning (FL) enables collaborative training over distributed multimedia data but suffers acutely from statistical heterogeneity and communication constraints, especially when clients deploy large models. Classic parameter-averaging methods such as FedAvg transmit full model weights and can diverge under nonindependent and identically distributed (non-IID) data. We propose KTA v2, a prediction-space knowledge trading market for FL. Each round, clients locally train on their private data, then share only logits on a small public reference set. The server constructs a client--client similarity graph in prediction space, combines it with reference-set accuracy to form per-client teacher ensembles, and sends back personalized soft targets for a second-stage distillation update. This two-stage procedure can be interpreted as approximate block-coordinate descent on a unified objective with prediction-space regularization. Experiments on FEMNIST, CIFAR-10 and AG News show that, under comparable or much lower communication budgets, KTA v2 consistently outperforms a local-only baseline and strong parameter-based methods (FedAvg, FedProx), and substantially improves over a FedMD-style global teacher. On CIFAR-10 with ResNet-18, KTA v2 reaches 57.7\% test accuracy using $\approx$1/1100 of FedAvg's communication, while on AG News it attains 89.3\% accuracy with $\approx$1/300 of FedAvg's traffic.
\end{abstract}

\section*{Keywords}
Federated learning, knowledge distillation, communication efficiency, non-IID data, multimedia learning.

\section{Introduction}
Many multimedia applications---mobile photo search, on-device news categorization, cross-camera video understanding---naturally fit the federated learning (FL) paradigm, in which data remain on devices while a server aggregates locally trained models \cite{mcmahan2017communication}. Communication-efficient variants of FL such as FedAvg reduce uplink traffic by aggregating model updates, but still transmit dense parameter tensors whose size scales with model depth \cite{mcmahan2017communication}. As models for vision and language grow larger, communication rather than computation often becomes the bottleneck.

A second challenge is statistical heterogeneity: in realistic multimedia deployments, each device observes its own skewed distribution over content types, users and contexts, and parameter-averaging algorithms like FedAvg and FedProx \cite{li2020federated} may suffer from client drift and slow or unstable convergence under such non-IID data \cite{karimireddy2020scaffold}. Personalized FL methods \cite{fallah2020personalized,dinh2020personalized} address part of this issue by learning client-specific models, but most still rely on exchanging parameters or gradients.

Knowledge distillation (KD) has emerged as a flexible mechanism for model compression and knowledge transfer. A growing body of work explores KD inside FL---FedMD \cite{li2019fedmd}, DS-FL \cite{itahara2023distillation}, data-free KD for heterogeneous FL \cite{zhu2021data} and feature-distillation approaches such as FedFed \cite{yang2023fedfed}---suggesting a move from parameter-space to prediction-space aggregation. Recent surveys systematically review this landscape and highlight persistent challenges around heterogeneity, privacy and communication \cite{qin2024knowledge,moslemi2024survey,salman2025knowledge}.

This paper proposes KTA v2, a prediction-space knowledge trading market designed for communication-efficient and personalized FL in multimedia tasks. We target the setting where a modest public reference set exists (e.g., unlabeled or weakly labeled multimedia content), and clients may use heterogeneous models and experience heavy label skew.

Our key idea is to build, each round, a knowledge market over client predictions on the reference set. Rather than aggregating parameters, the server measures pairwise similarity between clients in prediction space, weighs them by reference-set accuracy, and constructs for each client a personalized teacher ensemble whose soft labels are distilled back into the client. This yields several advantages:
\begin{itemize}
    \item communication complexity that scales with the prediction tensor size rather than model parameters;
    \item per-client teacher distributions that adapt to local data, unlike global teachers in FedMD-style methods;
    \item a natural interpretation as prediction-space regularization that mitigates cross-client drift.
\end{itemize}

We validate KTA v2 on three representative multimedia tasks---handwritten character recognition (FEMNIST), natural image classification (CIFAR-10) and news topic classification (AG News)---using both small CNNs and a larger ResNet-18 backbone. We compare against Local-only, FedAvg, FedProx \cite{li2020federated}, and a FedMD baseline \cite{li2019fedmd}, and relate our prediction-space regularization to SCAFFOLD-style insights \cite{karimireddy2020scaffold}.

Our main contributions are:
\begin{itemize}
    \item \textbf{Prediction-space knowledge market:} We introduce a framework that aggregates client predictions on a shared reference set via similarity--accuracy weighting, yielding personalized teacher ensembles per client rather than a single global teacher.
    \item \textbf{Unified objective and theoretical insight:} We show that KTA v2's two-stage training approximates block-coordinate descent on a single objective that combines local supervised loss with prediction-space regularization, and we discuss how this mitigates client drift at the level of logits rather than gradients.
    \item \textbf{Communication-efficient multimedia FL:} On vision and text benchmarks, KTA v2 consistently delivers competitive or superior accuracy under aggressive communication budgets, and can significantly outperform FedAvg for large models.
    \item \textbf{Systematic comparison with distillation-based FL:} We implement a FedMD baseline on CIFAR-10 and show that KTA v2 achieves substantially higher accuracy at similar communication cost, highlighting the benefit of per-client markets over global teachers.
\end{itemize}
\section{Related Work}
\subsection{Parameter-Based Federated Learning}
FedAvg remains the canonical FL algorithm, aggregating client-side stochastic gradient descent (SGD) updates by weighted averaging of model parameters \cite{mcmahan2017communication}. Numerous variants address heterogeneity and client drift. FedProx introduces a proximal term that keeps local models close to the global model, together with convergence analysis under heterogeneous objectives \cite{li2020federated}. SCAFFOLD uses control variates to correct for client drift and enjoys communication-efficient convergence guarantees \cite{karimireddy2020scaffold}. Additional work explores adaptive optimizers, variance reduction and client sampling strategies \cite{zhang2021semi}.

While powerful, these methods all transmit dense parameter tensors or gradients each round, making communication cost grow with model size. For multimedia applications where ResNet and transformer architectures are common, this can be prohibitive.

\subsection{Prediction-Based and Distillation-Based FL}
Knowledge distillation transfers information between models via soft predictions \cite{moslemi2024survey}. In FL, prediction-based methods aim to exchange logits or features instead of parameters, often leveraging public unlabeled data \cite{qin2024knowledge,salman2025knowledge}.

FedMD \cite{li2019fedmd} is a seminal heterogeneous FL framework: clients train local models on private data, then upload logits on a public dataset to form a global teacher (an ensemble of averaged predictions) which is distilled back to clients. DS-FL \cite{itahara2023distillation} extends this idea to semi-supervised settings, combining unlabeled public data with distillation to reduce communication. Zhu et al. propose data-free KD for FL, where the server learns a generator to synthesize inputs for distillation without access to real data \cite{zhu2021data}. More recent work, such as FedFed \cite{yang2023fedfed}, blends feature-level and logit-level sharing to mitigate heterogeneity.

These approaches primarily focus on constructing one or several global teachers and often treat all clients symmetrically when aggregating predictions, rarely exploiting pairwise prediction similarity to personalize the ensemble.

\subsection{Personalized Federated Learning}
Personalized FL (pFL) aims to tailor models to each client's data distribution. Per-FedAvg formulates pFL as a meta-learning problem, learning an initialization that can be quickly adapted per client \cite{fallah2020personalized}. pFedMe uses Moreau envelopes to decouple global and personalized optimization via bi-level formulations \cite{dinh2020personalized}. A rich line of work explores algorithmic and theoretical aspects of pFL, as surveyed by Sabah et al. \cite{sabah2024model} and Dinh \cite{dinh2024personalized}.

Many pFL methods still operate in parameter space and maintain separate local models at each client. Prediction-space approaches like FedMD can accommodate heterogeneous architectures \cite{li2019fedmd}, but typically still rely on global teachers. Our work instead combines prediction-space aggregation with per-client similarity--accuracy markets to yield lightweight personalization.
\section{Prediction-Space Knowledge Trading Market (KTA v2)}
\subsection{Problem Setup}
We consider $C$ clients, indexed by $i=1,\dots,C$. Client $i$ holds private labeled data $\mathcal{D}_i$ and maintains a local model $f_i(x;\theta_i)$. The goal is to collaboratively improve all clients via FL without sharing raw data.

We assume access to a small reference set $\mathcal{D}_{\mathrm{ref}} = \{(x_r,y_r)\}_{r=1}^{N_{\mathrm{ref}}}$ drawn from the same domain as the clients' data. In practice, $\mathcal{D}_{\mathrm{ref}}$ may be a held-out subset of public multimedia data or weakly labeled content. All methods, including baselines, may use $\mathcal{D}_{\mathrm{ref}}$ for validation; only prediction-based methods exchange logits on $\mathcal{D}_{\mathrm{ref}}$.

Each round $t$ proceeds in two stages:
\begin{enumerate}
    \item \textbf{Local supervised update:} Each client $i$ runs $E$ steps of SGD on $\mathcal{D}_i$ starting from its current $\theta_i$.
    \item \textbf{Knowledge market distillation:} Clients evaluate their models on $\mathcal{D}_{\mathrm{ref}}$, send predictions to the server; the server constructs per-client teachers based on a similarity--accuracy market; clients then run $E_{\mathrm{distill}}$ steps of distillation on $\mathcal{D}_{\mathrm{ref}}$ using these teachers.
\end{enumerate}
We next detail the market mechanism.

\subsection{Constructing the Knowledge Market}
Let $Z_c \in \mathbb{R}^{N_{\mathrm{ref}}\times K}$ denote the logits of client $c$ on the reference set with $K$ classes, and let $p_c(\cdot\mid x_r)$ be the corresponding softmax probability.

\textbf{Prediction-space similarity:} We flatten each client's prediction tensor to a vector $\tilde{z}_c = \mathrm{normalize}(\mathrm{vec}(Z_c)) \in \mathbb{R}^{N_{\mathrm{ref}}K}$ and define cosine similarity $S_{ij} = \tilde{z}_i^\top \tilde{z}_j$.

\textbf{Reference accuracy:} For supervised tasks we compute each client $j$'s accuracy on $\mathcal{D}_{\mathrm{ref}}$, $\alpha_j = \frac{1}{N_{\mathrm{ref}}}\sum_r \mathbf{1}\big[\arg\max_k p_j(k\mid x_r) = y_r\big]$, as a coarse quality estimate.

\textbf{Similarity--accuracy weights:} For a target client $i$, we choose a neighbor set $\mathcal{N}(i) \subseteq \{1,\dots,C\}\setminus\{i\}$. In the basic configuration we use $k$-nearest neighbors in $S$ (top-$k$ similarities) with $k=5$; a full-neighbor variant uses all other clients. For $j \in \mathcal{N}(i)$, define an unnormalized weight
\begin{equation}
    \tilde{w}_{ij} = \max(S_{ij},0)\cdot \max(\alpha_j,\varepsilon),
\end{equation}
where $\varepsilon>0$ avoids degenerate zero accuracy. The normalized weight is
\begin{equation}
    w_{ij} = \frac{\tilde{w}_{ij}}{\sum_{\ell\in\mathcal{N}(i)} \tilde{w}_{i\ell}}.
\end{equation}
A uniform ablation simply sets $w_{ij}=1/|\mathcal{N}(i)|$.

\textbf{Per-client teacher ensemble:} The teacher distribution for client $i$ on $x_r$ is the weighted ensemble
\begin{equation}
    q_i(\cdot\mid x_r) = \sum_{j\in\mathcal{N}(i)} w_{ij}\cdot \mathrm{softmax}\big(Z_j(r,:)/T\big),
\end{equation}
where $T$ is a temperature hyperparameter.

Intuitively, each client buys knowledge from ``nearby and accurate'' peers: prediction-similar clients with high reference accuracy contribute more to its teacher, while dissimilar or unreliable clients are downweighted. This yields personalized teachers across clients, in contrast to the single global teacher of FedMD \cite{li2019fedmd}.
\subsection{Two-Stage Training and Unified Objective}
For client $i$, KTA v2's two-stage training can be viewed as approximate minimization of a unified objective that combines local supervision and prediction-space regularization.

Let $\ell(f_i(x;\theta_i),y)$ be the standard cross-entropy loss on labeled data. Given fixed teacher distributions $q_i(\cdot\mid x)$ on $\mathcal{D}_{\mathrm{ref}}$, define
\begin{align}
    L_i(\theta_i) = &\,(1-\lambda)\,\mathbb{E}_{(x,y)\sim\mathcal{D}_i}[\ell(f_i(x;\theta_i),y)] \nonumber\\
    &+ \lambda T^2\, \mathbb{E}_{x\sim \mathcal{D}_{\mathrm{ref}}}\big[\mathrm{KL}\big(p_i(\cdot\mid x;\theta_i)\,\|\, q_i(\cdot\mid x)\big)\big],
    \label{eq:local_obj}
\end{align}
where $p_i(\cdot\mid x;\theta_i)$ is the softmax of client $i$'s logits and $\lambda\in[0,1]$ controls distillation strength.

Aggregating over all clients with weights proportional to their data sizes gives a global objective
\begin{equation}
    \tilde{F}(\theta_1,\dots,\theta_C) = \sum_i \frac{|\mathcal{D}_i|}{N_{\mathrm{total}}}\, L_i(\theta_i).
    \label{eq:global_obj}
\end{equation}

Within each FL round, KTA v2:
\begin{itemize}
    \item performs several gradient steps on the first term of \eqref{eq:local_obj} using local batches from $\mathcal{D}_i$;
    \item updates $q_i$ by recomputing teacher distributions from fresh predictions on $\mathcal{D}_{\mathrm{ref}}$;
    \item performs several gradient steps on the second term of \eqref{eq:local_obj} using reference-set batches.
\end{itemize}
Under small learning rates and fixed teachers within a round, this two-stage procedure approximates block-coordinate descent on $\tilde{F}$: the local phase updates $\theta_i$ along the supervised component, while the distillation phase pulls $p_i$ toward the market-consensus $q_i$ on $\mathcal{D}_{\mathrm{ref}}$. Unlike parameter-based corrections such as SCAFFOLD \cite{karimireddy2020scaffold}, the regularization operates directly in prediction space, which is invariant to model architecture as long as outputs share the same label space.

\subsection{Comparison to FedMD}
FedMD \cite{li2019fedmd} also shares predictions on a public dataset, but its teacher construction differs fundamentally from KTA v2:
\begin{itemize}
    \item FedMD averages all client predictions uniformly to form a single global teacher $q_{\mathrm{global}}(\cdot\mid x)$, whereas KTA v2 builds per-client teachers $q_i(\cdot\mid x)$ using similarity--accuracy weighting.
    \item FedMD does not exploit client-to-client similarity or reference accuracy; all clients receive identical distillation signals regardless of their local data.
    \item KTA v2's teacher selection reduces to FedMD when $\mathcal{N}(i)$ includes all clients and $w_{ij}=1/C$, providing a direct conceptual link.
\end{itemize}
Experiments on CIFAR-10 show that this difference is crucial: under the same communication budget ($\approx$8 MB of logits per round), KTA v2 substantially outperforms the FedMD baseline (Table~\ref{tab:key-comm}).
\subsection{Algorithm Outline}
For completeness, we summarize KTA v2's server and client procedures.

\textbf{Server (per round):}
\begin{enumerate}
    \item Collect $\mathcal{D}_{\mathrm{ref}}$ predictions $Z_c$ from participating clients.
    \item Compute $\tilde{z}_c$, similarity matrix $S$ and reference accuracies $\alpha_c$.
    \item For each client $i$:
    \begin{enumerate}
        \item choose neighbor set $\mathcal{N}(i)$;
        \item compute weights $w_{ij}$ from $S$ and $\alpha$;
        \item form $q_i(\cdot\mid x_r)$ by weighted ensemble over neighbors' softmax predictions.
    \end{enumerate}
    \item Send the corresponding teacher logits or soft labels for $q_i$ to each client $i$.
\end{enumerate}

\textbf{Client $i$ (per round):}
\begin{enumerate}
    \item Initialize $\theta_i$ from the previous round; run $E$ epochs of supervised training on local data $\mathcal{D}_i$.
    \item Receive teacher distributions $q_i(\cdot\mid x)$ from the server.
    \item Run $E_{\mathrm{distill}}$ epochs of mixed supervised+distillation training on $\mathcal{D}_{\mathrm{ref}}$, minimizing \eqref{eq:local_obj}.
\end{enumerate}

In all experiments, we also employ a BN-safe strategy: if a batch has size $\leq 1$, the corresponding update is skipped to avoid unstable Batch Normalization statistics, which only affects a small fraction of updates in our non-IID partitions.

\section{Theoretical Insights}
\subsection{Market Regularization as Consensus in Prediction Space}
Consider the logits vector $z_i(x)$ produced by client $i$ on an input $x\in \mathcal{D}_{\mathrm{ref}}$, and let $\bar{z}_i(x)$ be the logit implied by the market teacher $q_i$. The KL term in \eqref{eq:local_obj} induces gradients that nudge $z_i(x)$ towards $\bar{z}_i(x)$. For a fixed market graph and small learning rate $\eta$, a single distillation update step on a batch from $\mathcal{D}_{\mathrm{ref}}$ can be approximated as
\begin{equation}
    z_i \leftarrow z_i - \eta \nabla_{z_i} \mathrm{KL}(p_i\|q_i) \approx (1-\eta\mu)z_i + \eta \mu \sum_{j\in\mathcal{N}(i)} w_{ij} z_j,
    \label{eq:consensus}
\end{equation}
for some scalar $\mu$ depending on $T$ and local curvature. This has the same structure as a consensus update on a weighted graph, where each node moves towards a convex combination of its neighbors' logits.

Compared with parameter-space methods such as SCAFFOLD \cite{karimireddy2020scaffold}, which enforce consistency of gradients or parameters, KTA v2 enforces prediction-space consistency on $\mathcal{D}_{\mathrm{ref}}$. This is particularly important for multimedia models where architectures may differ across clients: predictions are always comparable as long as they share a label vocabulary.

\subsection{Client Drift and Heterogeneity}
Non-IID label and feature distributions cause local optima of $L_i$ to differ across clients, leading to drift in parameter-based methods \cite{karimireddy2020scaffold}. In KTA v2, the consensus pressure from \eqref{eq:consensus} acts as a form of prediction drift reduction: clients with similar predictive behavior and high reference accuracy exert stronger influence on each other, while outliers contribute less through the similarity--accuracy weights.

On CIFAR-10, we empirically observe that the variance of per-client accuracy across rounds is lower under KTA v2 than under FedAvg at the same communication budget (Section~\ref{sec:results}), supporting this view.
\section{Experimental Setup}
\subsection{Datasets and Models}
We evaluate on three benchmarks representative of multimedia tasks:
\begin{itemize}
    \item \textbf{FEMNIST:} 62-class handwritten characters and digits derived from EMNIST, partitioned by writer into 20 clients (grayscale $28\times 28$ images). We use a small CNN with three convolutional blocks and BatchNorm.
    \item \textbf{CIFAR-10:} 10-class natural image classification with 50k training and 10k test examples (RGB $32\times 32$). We consider (i) a SimpleCNN comparable in size to prior FL work; (ii) a ResNet-18 backbone to emulate realistic multimedia models.
    \item \textbf{AG News:} 4-class news topic classification. We build a vocabulary of 20k tokens and use an embedding + mean-pooling + linear classifier as a lightweight text model.
\end{itemize}
For each dataset we hold out a small labeled reference set $\mathcal{D}_{\mathrm{ref}}$ of size $N_{\mathrm{ref}}=2000$ drawn from the training distribution. All methods see the same reference data; only KTA v2 and FedMD exchange predictions on it.

\subsection{Federated Partitioning and Non-IIDness}
We construct non-IID client partitions using a Dirichlet distribution over labels with concentration parameter $\alpha$. For a given dataset and client count $C$, each class's samples are split across clients according to a Dirichlet($\alpha$) draw, producing label-skewed partitions. We use:
\begin{itemize}
    \item FEMNIST: $C=20$, $\alpha=0.5$;
    \item CIFAR-10: $C=10$, $\alpha \in \{0.1,0.5,1.0\}$ (for the non-IID sweep in Fig.~\ref{fig:cifar-alpha})
    \item AG News: $C=10$, $\alpha=0.5$.
\end{itemize}
Unless otherwise noted, we report results for $\alpha=0.5$ as the main setting.

\subsection{Baselines and Hyperparameters}
We compare:
\begin{itemize}
    \item \textbf{Local:} clients train local models independently with no communication.
    \item \textbf{FedAvg:} standard parameter averaging \cite{mcmahan2017communication}.
    \item \textbf{FedProx:} FedAvg with proximal term $\mu=0.01$ \cite{li2020federated}.
    \item \textbf{FedMD:} simplified heterogeneous FL via model distillation \cite{li2019fedmd}, in which a global teacher is formed by uniformly averaging client predictions on $\mathcal{D}_{\mathrm{ref}}$ and distilled back to clients.
    \item \textbf{KTA v2:} our prediction-space knowledge market.
\end{itemize}
For all methods we use Adam with learning rate $10^{-3}$, batch size 64 and 10 global rounds (5 rounds for ResNet-18 due to cost). Each round consists of one local epoch $E=1$; KTA v2 adds five distillation epochs per round on $\mathcal{D}_{\mathrm{ref}}$. For fair comparison, FedMD uses the same reference set size and number of rounds as KTA v2 on CIFAR-10.

We measure communication cost in megabytes (MB) of float32 values sent over the uplink and downlink:
\begin{itemize}
    \item parameter-based methods: per round, each participating client uploads and downloads full model parameters (4 bytes per scalar);
    \item prediction-based methods: per round, each participating client uploads logits on $\mathcal{D}_{\mathrm{ref}}$, and receives soft labels or logits of the same size.
\end{itemize}
All reported communication numbers are cumulative over all rounds.

\subsection{Metrics and Reporting}
We log per-client and global test accuracy and loss at each round. For main configurations we run three random seeds and report mean $\pm$ standard deviation over seeds. Tables~\ref{tab:overall} and \ref{tab:key-comm} summarize last-round results.
\section{Results and Discussion}
\label{sec:results}
\subsection{Overall Accuracy and Communication}
Table~\ref{tab:overall} reports global test accuracy and total communication on the three datasets for Local, FedAvg, FedProx, FedMD (CIFAR-10) and KTA v2.

\begin{table}[t]
    \centering
    \caption{Overall accuracy and communication (mean $\pm$ std over three seeds).}
    \label{tab:overall}
    \scriptsize
    \setlength{\tabcolsep}{2pt}
    \begin{tabular}{l l c c}
        \toprule
        Dataset & Method & Acc. (\%) & Comm. (MB) \\
        \midrule
        FEMNIST  & Local   & 45.2 $\pm$ 1.2 & 0.0   \\
        FEMNIST  & FedAvg  & 74.3 $\pm$ 0.1 & 154.3 \\
        FEMNIST  & FedProx & 74.1 $\pm$ 0.4 & 154.3 \\
        FEMNIST  & KTA v2  & 74.5 $\pm$ 1.9 & 94.6  \\
        CIFAR-10 & Local   & 37.4 $\pm$ 3.4 & 0.0   \\
        CIFAR-10 & FedAvg  & 57.1 $\pm$ 1.3 & 72.5  \\
        CIFAR-10 & FedProx & 57.8 $\pm$ 0.0 & 72.5  \\
        CIFAR-10 & FedMD   & 38.0 $\pm$ 3.2 & 8.0   \\
        CIFAR-10 & KTA v2  & 49.3 $\pm$ 8.0 & 7.6   \\
        AG News  & Local   & 66.8 $\pm$ 5.6 & 0.0   \\
        AG News  & FedAvg  & 87.0 $\pm$ 0.3 & 976.8 \\
        AG News  & FedProx & 86.9 $\pm$ 0.0 & 976.8 \\
        AG News  & KTA v2  & 89.3 $\pm$ 0.3 & 3.1   \\
        \bottomrule
    \end{tabular}
    \vspace{-1mm}
\end{table}

All CIFAR-10 entries use a SimpleCNN backbone unless otherwise noted; the ResNet-18 case appears in Table~\ref{tab:key-comm}.

On FEMNIST, KTA v2 matches or slightly exceeds FedAvg/FedProx in accuracy (74.5\% vs. 74.3\% / 74.1\%) while reducing communication by $\approx$39\%. On CIFAR-10 with SimpleCNN, KTA v2 attains 49.3\% accuracy using only 7.6 MB, compared with 57.1--57.8\% for FedAvg/FedProx at $\approx$72.5 MB; this is a $\approx$9.5$\times$ reduction in communication for a moderate drop in accuracy. Importantly, KTA v2 substantially outperforms FedMD, which only reaches 38.0\% at roughly the same communication budget.

For AG News, KTA v2 not only compresses communication by $\approx$300$\times$ relative to FedAvg/FedProx (3.1 MB vs. 976.8 MB), but also improves accuracy to 89.3\% from 87.0--86.9\%. This shows that, in text-based multimedia tasks with strong heterogeneity, prediction-space markets can be strictly better than parameter-based methods under realistic budgets.

\subsection{Trade-offs on CIFAR-10 and Comparison with FedMD}
\begin{figure}[!t]
    \centering
    \includegraphics[width=\linewidth]{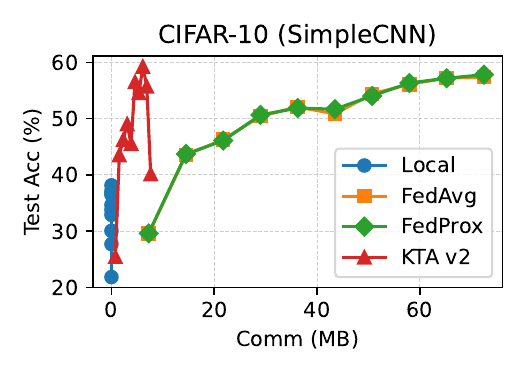}
    \caption{CIFAR-10 (SimpleCNN) accuracy versus communication for Local, FedAvg, FedProx and KTA v2 at $\alpha=0.5$; KTA v2 stays within 0--8 MB.}
    \label{fig:cifar-main}
\end{figure}

Figure~\ref{fig:cifar-main} plots CIFAR-10 (SimpleCNN) accuracy versus communication across FL rounds for Local, FedAvg, FedProx and KTA v2. Local training improves with zero communication but saturates around 37--39\%. FedAvg and FedProx exhibit smooth accuracy growth as communication increases, eventually approaching 58\% by $\sim$70 MB. KTA v2 operates in a very low-communication regime (0--8 MB), yet achieves 45--60\% accuracy during the training trajectory; its final point at 7.6 MB dominates Local and FedMD, and provides a compelling trade-off against FedAvg/FedProx.

Table~\ref{tab:key-comm} further summarizes these communication-efficient configurations. Comparing the CIFAR-10 / SimpleCNN entries, KTA v2 improves accuracy by more than 11 percentage points over FedMD at essentially the same communication cost.
\begin{table}[t]
    \centering
    \caption{Key communication-efficient configurations.}
    \label{tab:key-comm}
    \scriptsize
    \setlength{\tabcolsep}{2pt}
    \begin{tabular}{l l c c}
        \toprule
        Case & Method & Acc. (\%) & Comm. (MB) \\
        \midrule
        CIFAR-10 / ResNet-18 & FedAvg        & 42.1         & 4265.5 \\
        CIFAR-10 / ResNet-18 & KTA v2        & 57.7         & 3.8    \\
        AG News              & Local         & 66.8 $\pm$ 5.6 & 0.0   \\
        AG News              & FedAvg (main) & 87.0 $\pm$ 0.3 & 976.8 \\
        AG News              & FedAvg (low)  & 53.3 $\pm$ 0.0 & 97.7  \\
        AG News              & KTA v2 (main) & 89.3 $\pm$ 0.3 & 3.1   \\
        CIFAR-10 / SimpleCNN & FedMD         & 38.0 $\pm$ 3.2 & 8.0   \\
        CIFAR-10 / SimpleCNN & KTA v2        & 49.3 $\pm$ 8.0 & 7.6   \\
        \bottomrule
    \end{tabular}
    \vspace{-1mm}
\end{table}

These comparisons underscore that per-client similarity--accuracy weighting is more effective than a uniform global teacher in non-IID image classification.

\subsection{Large-Model Regime: CIFAR-10 + ResNet-18}
\begin{figure}[!t]
    \centering
    \includegraphics[width=\linewidth]{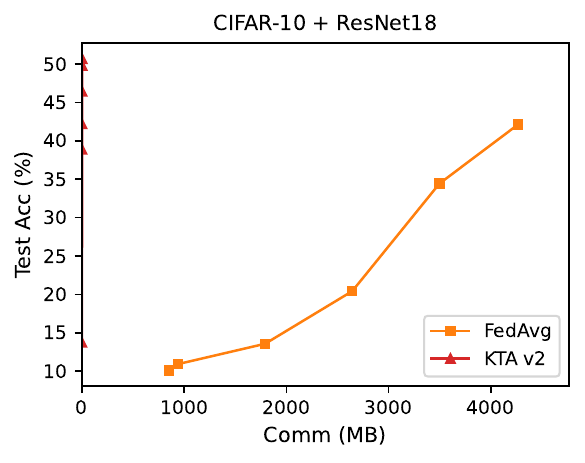}
    \caption{CIFAR-10 + ResNet-18 communication/accuracy trajectory. KTA v2 reaches 57.7\% with $\approx$3.8 MB, while FedAvg attains 42.1\% at $\approx$4265.5 MB.}
    \label{fig:resnet}
\end{figure}

Figure~\ref{fig:resnet} shows the communication--accuracy trajectory on CIFAR-10 with ResNet-18 for FedAvg and KTA v2. Each FedAvg point corresponds to an additional global round, while KTA v2 operates in a narrow low-communication band.

KTA v2 reaches 57.7\% accuracy with only 3.8 MB of cumulative communication, whereas FedAvg requires 4.3 GB ($\approx$4265.5 MB) to reach 42.1\%. This represents roughly 1118$\times$ communication reduction together with a substantial accuracy gain. The gap arises because parameter size grows with model depth, whereas KTA v2's prediction tensors remain fixed by $N_{\mathrm{ref}}$ and $K$.

\subsection{Non-IID Robustness}
\begin{figure}[!t]
    \centering
    \includegraphics[width=\linewidth]{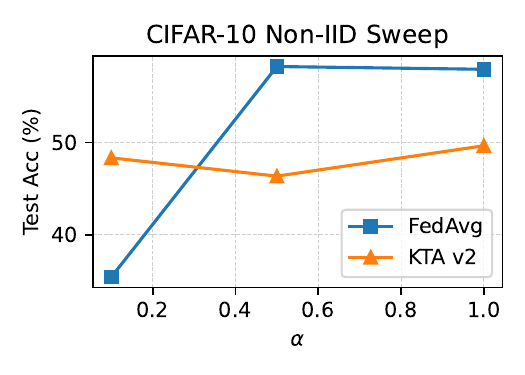}
    \caption{CIFAR-10 non-IID sweep (SimpleCNN). Test accuracy after 10 rounds under Dirichlet $\alpha$; smaller $\alpha$ means stronger label skew.}
    \label{fig:cifar-alpha}
\end{figure}

Figure~\ref{fig:cifar-alpha} plots final test accuracy versus Dirichlet $\alpha$ for CIFAR-10 (SimpleCNN), comparing FedAvg and KTA v2. Smaller $\alpha$ corresponds to stronger label skew.

At $\alpha=0.1$ (severe non-IID), FedAvg drops to around 36\% accuracy, while KTA v2 maintains $\approx$49\%. As $\alpha$ increases to 0.5 and 1.0, FedAvg improves significantly, eventually surpassing KTA v2 in accuracy under its much higher communication budget. This pattern suggests that KTA v2 is particularly attractive in highly heterogeneous regimes where parameter averaging struggles, while still remaining competitive under milder heterogeneity.

\subsection{BN-Safe Training and Reference Set Discussion}
In all experiments we enable a BN-safe rule: any batch with size $\leq 1$ is skipped to avoid unstable BatchNorm statistics. We empirically find that this affects less than 3\% of updates in our partitions and has negligible impact on data utilization. Disabling it leads to occasional divergence on CIFAR-10 with ResNet-18, underscoring the importance of this simple engineering safeguard in federated regimes with small per-client datasets.

Regarding reference set fairness, we construct $\mathcal{D}_{\mathrm{ref}}$ by uniformly sampling from the training distribution. All methods may use $\mathcal{D}_{\mathrm{ref}}$ for validation or early stopping. Only prediction-based methods (FedMD, KTA v2) additionally use it for distillation; however, they do not see extra labels beyond what FedAvg/FedProx already use, since $\mathcal{D}_{\mathrm{ref}}$ is labeled training data. In practical deployments, $\mathcal{D}_{\mathrm{ref}}$ could be an unlabeled or weakly labeled public multimedia corpus, combined with pseudo-labels or self-supervision.

\subsection{Discussion}
Overall, KTA v2 is most attractive in communication-limited multimedia FL and large-model regimes, where prediction-space sharing decouples traffic from model size while still providing an implicit form of personalization via per-client teacher ensembles.
\section{Conclusion}
We presented KTA v2, a prediction-space knowledge trading market for communication-efficient federated learning on multimedia tasks. By exchanging only logits on a public reference set and constructing per-client teacher ensembles using similarity--accuracy weighting, KTA v2 achieves strong performance under non-IID data and aggressive communication budgets. A unified objective interpretation shows that KTA v2 can be seen as adding prediction-space consensus regularization, helping mitigate client drift without restricting model architectures.

Experiments on FEMNIST, CIFAR-10 (SimpleCNN and ResNet-18) and AG News demonstrate that KTA v2 is competitive with or superior to parameter-based baselines and a FedMD-style global teacher, especially in large-model and highly heterogeneous regimes. Future work includes extending the market to multi-modal reference sets (e.g., image--text pairs, video clips), integrating stronger privacy guarantees for logit sharing, and exploring adaptive market graphs that dynamically adjust to evolving client populations.

\end{document}